\documentclass[runningheads]{llncs}

\usepackage{graphicx}

\usepackage{amsmath,amssymb}
\usepackage{color}
\usepackage{multirow}
\usepackage{subfig}
\usepackage{booktabs}
\usepackage{verbatim}

\begin{document}
\title{Language Guided Fashion Image Manipulation with Feature-wise Transformations\thanks{This is an extended version of a paper with the same title that has been accepted for presentation at the First Workshop on Computer Vision For Fashion, Art and Design at ECCV 2018. This research was supported in part by TUBITAK with award no 217E029. We would like to thank NVIDIA Corporation for the donation of a Quadro P5000 GPU.}}
\titlerunning{Language Guided Fashion Image Manipulation with FiLM}

\author{Mehmet G\"{u}nel \and
Erkut Erdem \and
Aykut Erdem}

\authorrunning{M. G\"{u}nel et al.}

\institute{Hacettepe University Computer Vision Lab\\Dept. of Computer Engineering, Hacettepe University, Ankara, Turkey\\ \email{\{n14327319,erkut,aykut\}@cs.hacettepe.edu.tr}}

\maketitle
\begin{abstract}
Developing techniques for editing an outfit image through natural sentences and accordingly generating new outfits has promising applications for art, fashion and design. However, it is considered as a certainly challenging task since image manipulation should be carried out only on the relevant parts of the image while keeping the remaining sections untouched. Moreover, this manipulation process should generate an image that is as realistic as possible. In this work, we propose FiLMedGAN, which leverages feature-wise linear modulation (FiLM) to relate and transform visual features with natural language representations without using extra spatial information. Our experiments demonstrate that this approach, when combined with skip connections and total variation regularization, produces more plausible results than the baseline work, and has a better localization capability when generating new outfits consistent with the target description.
\end{abstract}

\section{Introduction}
Language based image editing (LBIE)~\cite{chen2017language} is a recently proposed umbrella term which describes the task of transforming a source image based on natural language descriptions. A specific case of LBIE aims at modifying an outfit in an image using textual descriptions as target transformations~\cite{iccv2017fashiongan}, which has potential applications in art, fashion, shopping and design. However, this is a rather challenging problem mainly due to two reasons. A successful model should be able to (i) reflect the changes to the input image while preserving structural coherence (\textit{e.g.} body shape, pose, person identity), and (ii) understand and resolve the local changes in images according to only the relevant parts of textual description. While the former is about the image generation process, the latter is related to understanding the relations between the source image and the language description and it requires to disentangle semantics from both visual and textual modalities. In this respect, it shares some similarities with other integrated vision and language problems such as visual question answering (VQA).

\begin{figure} [!t]
\centering
\begin{tabular}{cc@{$\quad$}ccccccc}
\footnotesize{input image} & & \multicolumn{7}{c}{\footnotesize{generated outfit images based on different textual descriptions}}\\
\cline{1-1}
\cline{3-9}\\
\multirow{2}{*}{\parbox{2.025cm}{\vspace{-1.4cm} \includegraphics[height=2.70cm,width=2.025cm]{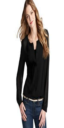}}} & &
\includegraphics[height=1.70cm,width=1.275cm]{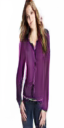} &
\includegraphics[height=1.70cm,width=1.275cm]{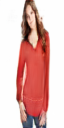} &
\includegraphics[height=1.70cm,width=1.275cm]{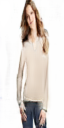} & 
\includegraphics[height=1.70cm,width=1.275cm]{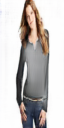} &
\includegraphics[height=1.70cm,width=1.275cm]{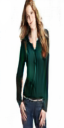} &
\includegraphics[height=1.70cm,width=1.275cm]{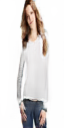} &
\includegraphics[height=1.70cm,width=1.275cm]{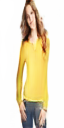}\\ & &
\includegraphics[height=1.70cm,width=1.275cm]{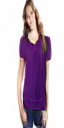} &
\includegraphics[height=1.70cm,width=1.275cm]{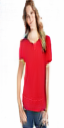} &
\includegraphics[height=1.70cm,width=1.275cm]{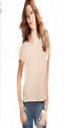} &
\includegraphics[height=1.70cm,width=1.275cm]{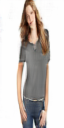} &
\includegraphics[height=1.70cm,width=1.275cm]{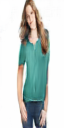} &
\includegraphics[height=1.70cm,width=1.275cm]{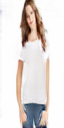} &
\includegraphics[height=1.70cm,width=1.275cm]{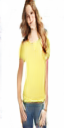} \\ 
 & &
\includegraphics[height=1.70cm,width=1.275cm]{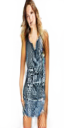} &
\includegraphics[height=1.70cm,width=1.275cm]{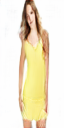} & 
\includegraphics[height=1.70cm,width=1.275cm]{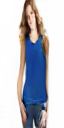} &
\includegraphics[height=1.70cm,width=1.275cm]{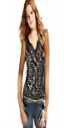} &
\includegraphics[height=1.70cm,width=1.275cm]{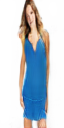} &
\includegraphics[height=1.70cm,width=1.275cm]{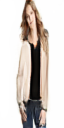} &
\includegraphics[height=1.70cm,width=1.275cm]{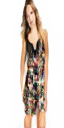}
\end{tabular}
\caption{Our model learns to manipulate fashion images via textual descriptions. Given an image (leftmost), it can synthesize new images with intended outfit changes such as ``\textit{the lady is wearing a multicolor sleeveless romper}'' (bottom rightmost).}
\label{fig:teaser}
\end{figure}

The main motivation of this paper comes from a recent conditioning mechanism known as Feature-wise Linear Modulation (FiLM), which has been initially proposed for solving complicated VQA tasks~\cite{perez:hal-01648685} and has been proven very useful. In this work, we propose a new conditional Generative Adversarial Network (GAN), which we name FiLMedGAN, which incorporates FiLM based feature transformations to better guide the manipulation process based on natural language descriptions. To increase the overall quality of the resulting images, our network architecture also employs skip connections \cite{RFB15a} and we additionally use total variation regularization \cite{Rudin1992} during training. We demonstrate that our proposed approach can synthesize and modify plausible outfit images \textit{without} a need to utilize extra spatial information like segmentation maps or body joints and pose guidance as commonly considered in the previous work~(see Fig.~\ref{fig:teaser}).

\section{Related Work}
Our model is based on Generative Adversarial Networks (GANs) \cite{NIPS2014_5423}. GANs have become one of the dominant methods to build generative models of complex, real-life data and many GAN variants have been proposed for a range of generation tasks. GANs can be formulated as a two-player game where a discriminator ($D$) and a generator ($G$) are trained in an alternating manner with an adversarial loss. Despite the difficulties in training \cite{Salimans:2016:ITT:3157096.3157346}, adversarial learning has been applied to numerous domains such as text-to-image synthesis \cite{pmlr-v48-reed16,Zhang_2017_ICCV,Han17stackgan2,zhang2018hdgan,Tao18attngan}, language based image editing \cite{DBLP:conf/iccv/DongYWG17,iccv2017fashiongan}, person image generation \cite{NIPS2017_6644,siarohin:hal-01761539,pumarola2018unsupervised,Lassner:GeneratingPeople:2017,ma2018disentangled,DBLP:conf/iccvw/JetchevB17} and texture synthesis~\cite{DBLP:journals/corr/JetchevBV16,pmlr-v70-bergmann17a,wang2018sftgan,Xian_2018_CVPR}.

The most relevant work to ours is by Dong \textit{et al.} \cite{DBLP:conf/iccv/DongYWG17} who first learn a visual-semantic text embedding from the image-text description pairs and then adversarially train a conditional generator network to perform LBIE on bird images from Caltech-200 \cite{WahCUB_200_2011} and flower images from Oxford-102 \cite{4756141}. Our model is built upon these ideas and indeed can be viewed as an improved version of that work. The details of this model and the extensions we propose in this paper will be described fully in Sec. \ref{network_arch}. 

In another related work, Zhu \textit{et al.} \cite{iccv2017fashiongan} performed LBIE on fashion images and proposed a model called FashionGAN. They emphasized structural coherence, which involves retaining body shape and pose, producing image parts that conform to given language description and enforcing coherent visibility of body parts. For that purpose, they proposed a two stage generator model, which also takes a human parsed segmentation map of the input image as complementary information. In the second stage, they generated target image conditioned on the segmentation map that is generated from the first stage together with language description. This differs significantly from our approach since we do not require any segmentation map or employ explicit spatial constraints, which is costly to obtain and might not always be available. We also believe that directly using segmentation maps in synthesizing the output might introduce some visual inconsistencies between the generated output and the actual input as the output is not generated in a holistic manner.

Other related works rather focus on image generation from text instead of directly manipulating images. Lassner \textit{et al.} \cite{Lassner:GeneratingPeople:2017} proposed a model (ClothNet) which is able to generate full body images of people with clothing conditioned on a specific pose, shape and color. CAGAN \cite{DBLP:conf/iccvw/JetchevB17} and VITON \cite{han2017viton} models take a person and a clothing image as inputs to dress up the person with the specified clothing item. The rest of the related work mostly concerns with the pose of humans and accordingly adds some spatial constraints \cite{NIPS2017_6644,ma2018disentangled,siarohin:hal-01761539,pumarola2018unsupervised,vunet2018,Zanfir_2018_CVPR}, which also differs from our work in this respect.

A popular approach to increase the quality of generated images is to incorporate attention mechanisms into GANs~\cite{DBLP:journals/corr/abs-1802-06454,Kastaniotis2018AttentionAwareGA,Tao18attngan,Chen2018AttentionGANFO,DBLP:journals/corr/abs-1805-08318}, which helps identifying the most relevant parts of images or features as needed. Ma \textit{et al.} \cite{DBLP:journals/corr/abs-1802-06454} proposed a deep attention encoder as a part of their model for instance level translation. Kastaniotis \textit{et al.} \cite{Kastaniotis2018AttentionAwareGA} used an attention mechanism in discriminator for generating better cell images. Xu \textit{et al.}~\cite{Tao18attngan} proposed attention driven multi stage refinement approach for text-to-image problems. Chen \textit{et al.}~\cite{Chen2018AttentionGANFO} proposed an extra attention network for object transfiguration . A recent work of Zhang \textit{et al.}~\cite{DBLP:journals/corr/abs-1805-08318} involves a self attention model for retaining global consistency in generated images. Compared to these previous works, we alternatively explore using FiLM transformations \cite{perez:hal-01648685} as a conditioning mechanism and exploit its implicit attention-like mechanism. Quite recently, FiLM has been investigated in a similar manner for colorization~\cite{N18-2120} and recovering textures~\cite{wang2018sftgan}. However, the former does not use GANs and the task explored in the latter is quite dissimilar to ours.

\section{Methodology}
\label{network_arch}
\begin{figure}[!t]
\centering
\subfloat[][]{\includegraphics[height=4.5cm]{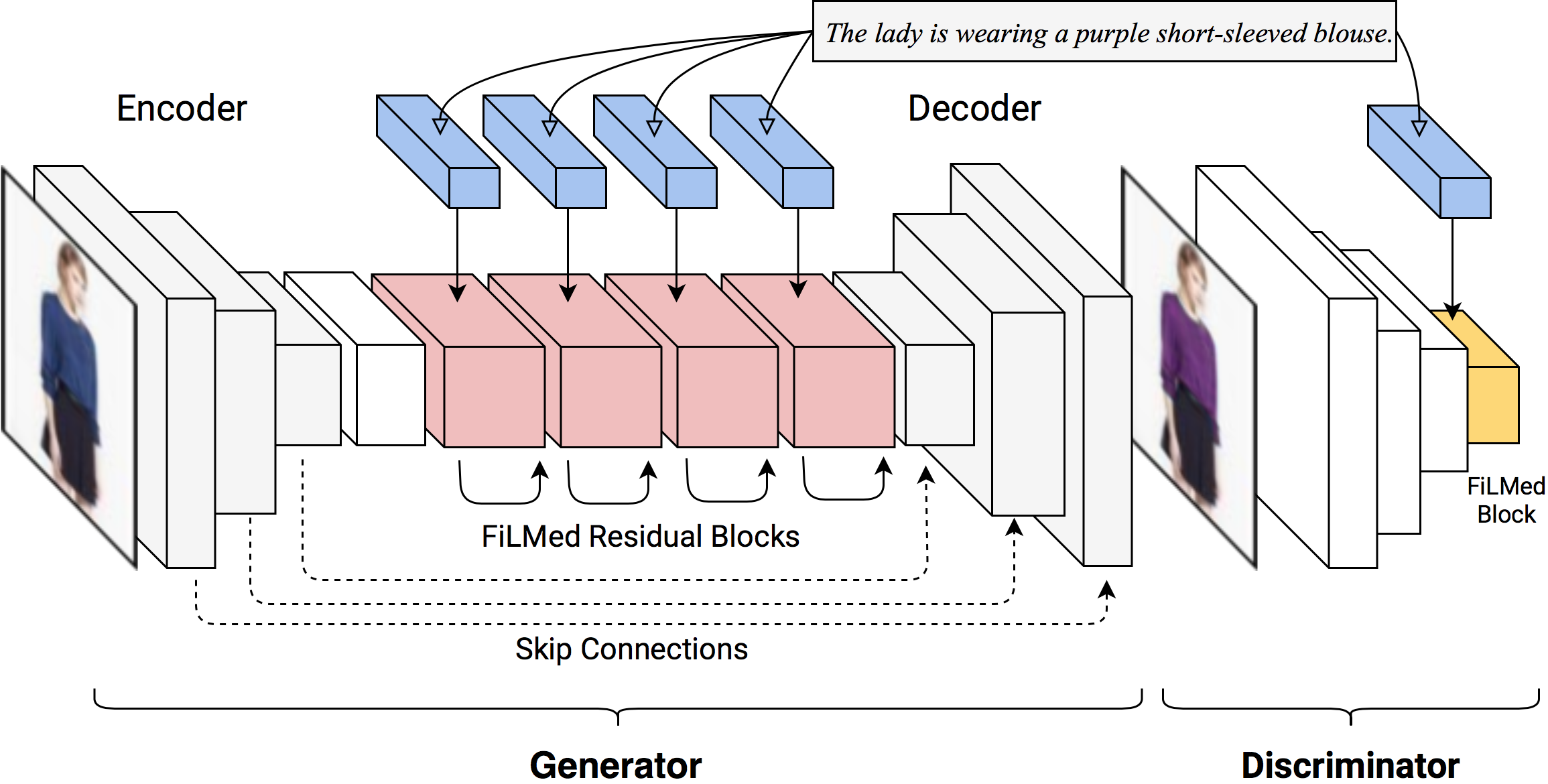}\label{fig:model}} \qquad
\subfloat[][]{\includegraphics[height=3cm]{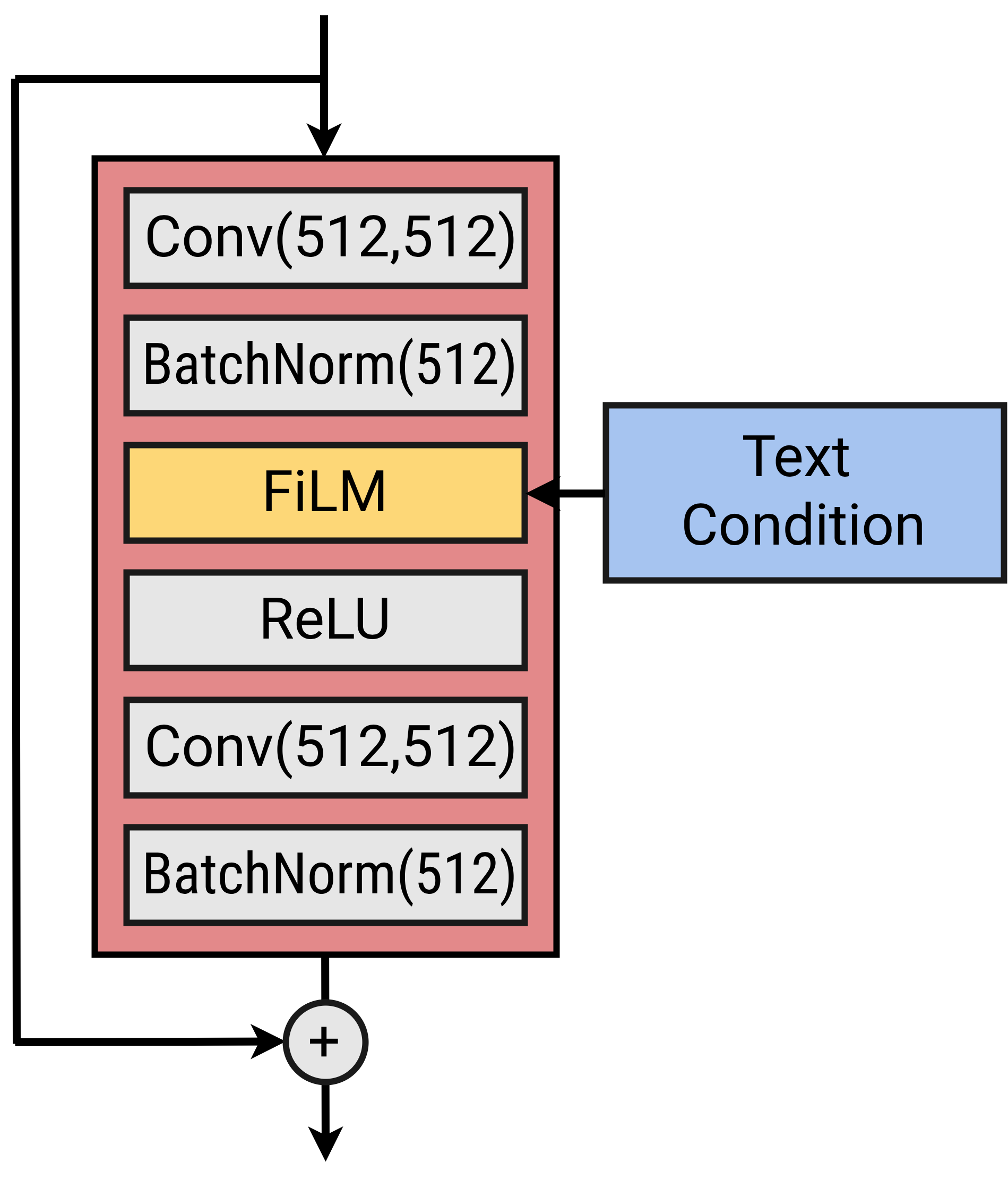}\label{fig:residualfilm}}
\caption{(a) The FiLMedGAN architecture, (b) Inside of a FiLMed residual block.}
\end{figure}

\textbf{Network Architecture.} Our network architecture is an improved version of the model suggested by Dong \textit{et al.} \cite{DBLP:conf/iccv/DongYWG17}, which is indeed inspired by \cite{Zhang_2017_ICCV} and \cite{Johnson2016Perceptual}. Fig.~\ref{fig:model} gives an overview of our improved architecture. The generator network in~\cite{DBLP:conf/iccv/DongYWG17} is made up of an encoder, a residual transformation unit and a decoder. The encoder and the decoder consists of 2D convolution layers together with several strides and nearest-neighbor up-samplings followed by ReLU activations and Batch Normalization (BN)~\cite{pmlr-v37-ioffe15}, except the first and the last layers respectively. We extend the encoder by including an extra 2D convolution and BN layers, which adds new features of dimensions $32\times 16\times 512$. Moreover, the feature maps of the encoder are concatenated to the corresponding decoder layers via symmetric skip connections. The residual transformation unit is made up of four residual blocks. We redesign this part by adding a FiLM block after the first BN layer. The architecture of our modified residual block can be seen in Fig.~\ref{fig:residualfilm}. The discriminator has also an encoder with a residual branch composed of convolution layers akin to the generator followed by a classifier layer. We modify it to process $128\times64$ images instead of $64\times64$ sized images by adding (2,1) strides to convolution layers. We also fuse together semantic embedding of textual input with encoder output using FiLM rather than a simple replication and concatenation, as done in~\cite{DBLP:conf/iccv/DongYWG17}.

To train our visual-semantic text embedding, we utilize an external pre-trained word embedding (fastText) \cite{bojanowski2017enriching} and early layers of pre-trained VGG-16 \cite{DBLP:journals/corr/SimonyanZ14a} network, as done in~\cite{DBLP:conf/iccv/DongYWG17}, where sentences are represented as the output of a GRU~\cite{cho-al-emnlp14} unit, and for training, we follow the same procedure in~\cite{DBLP:conf/iccv/DongYWG17} where we utilize a pairwise ranking loss.\\ 

\noindent\textbf{Improved Conditioning using FiLM.} Our inspiration comes from a recent work by Manjunatha \textit{et al.} \cite{N18-2120} on colorization of gray scale images with natural language descriptions who explore the use of FiLM to fuse textual representations with visual representations. Although qualitative results are not much different than those of simple concatenation, the authors reported that the activations of FiLM layer can emulate guided attention. This is in line with our own observations that it helps better localization while manipulating an image based on a textual description.

Mathematically speaking, a FiLM layer performs a feature-wise affine transform on visual features conditioned on textual information. Given $h$ as continuous vector representation of natural language description, we compute $\gamma_k$ and $\beta_k$ vectors as in Eqn.~\ref{filmmath1} where $W_\gamma$ and $W_\beta$ are parameters to be learned.

\begin{minipage}[t][0.8cm][t]{0.45\linewidth}
\begin{equation} 
 \gamma_k = W_\gamma h
 \quad
 \beta_k = W_\beta h
 \label{filmmath1}
\end{equation}
\end{minipage}
\begin{minipage}[t][0.8cm][t]{0.45\linewidth}
\begin{equation}
    Z^{\prime}_{k_{i,j}} = Z_{k_{i,j}} \circ \gamma_k + \beta_k
    \label{filmmath2}
\end{equation}
\end{minipage}

\noindent Here, $Z_k$ denotes a feature output and it is modulated as in Eqn.~\ref{filmmath2}, where $\circ$ is element-wise product and $k_{i,j}$ represents the spatial dimensions. For  implementation fusing vectors with concatenation might result an increase in parameter size of the network whereas FiLM is much more efficient.\\

\noindent\textbf{Regularization with Total Variation (TV).} In our FiLMed experiments, we come across with some artifacts and blur in some manipulated images. To overcome this issue, we additionally include total variation loss~\cite{Mahendran2015} as a regularization term to the loss function of the generator.
So, our final adversarial loss function that we use in our FiLMedGAN model becomes:
\begin{multline}
\mathcal{L}_D = \mathbb{E}_{(x,t)\sim p_{data}}logD(x, \varphi(t)) + 
\mathbb{E}_{(x,\hat{t})\sim p_{data}}log(1-D(x,\varphi(\hat{t})) + \\
\mathbb{E}_{(x,\bar{t})\sim p_{data}}log(1-D(G(x,\varphi(\bar{t})),\varphi(\bar{t})))
\end{multline}
\begin{equation}
    \mathcal{L}_G = \mathbb{E}_{(x,\bar{t})\sim p_{data}}log(D(G(x,\varphi(\bar{t})),\varphi(\bar{t}))) + \\ \lambda \mathcal{R}_{TV}(G(x,\varphi(\bar{t}))
\end{equation}
where $\varphi(t)$ stands for the matching text, $\varphi(\hat{t})$ represents a mismatching text and finally $\varphi(\bar{t})$ denotes a semantically relevant text~\cite{DBLP:conf/iccv/DongYWG17}.

\section{Experiments}

\textbf{Dataset.} In our experiments, we use Fashion Synthesis~\cite{iccv2017fashiongan} dataset, an extension of \cite{liu2016deepfashion}, which contains 78,979 images along with textual descriptions. It also provides gender, color, sleeve and category attributes as well as segmentation maps. We utilize the provided training (70,000) and test (8,979) splits and do not make use of the segmentation maps or the attributes during training.\\

\noindent
\textbf{Implementation Details and Training.} To train visual-semantic text embedding, we use the Adam optimizer \cite{DBLP:journals/corr/KingmaB14} with the parameters $\beta_1=0.9$, $\beta_2=0.999$, $\epsilon =10^8$ and a learning rate of $0.002$. We train our model with the batch size of $64$ for $200$ epochs. We set the pairwise ranking loss margin to~0.2, embedding dimension to 300 and max words to 25. Similarly, to train GAN model, we employ the Adam optimizer with a different parameter $\beta_1=0.5$, the rest is the same as visual-semantic text embedding training parameters. We also apply learning rate decay for 100 epochs with $\gamma=0.5$. We set the parameter $\lambda$  of the TV regularization term to $0.01$. All GAN models are trained for 125 epochs.\\

\begin{figure}[!t]
  \centering
  \subfloat[]{\label{ref_label1}\includegraphics[height=0.15\textwidth]{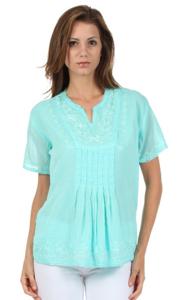}}
  \hspace{0.5cm}
	\subfloat[]{\label{ref_label2}\includegraphics[height=0.15\textwidth]{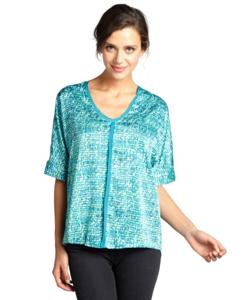}
	\includegraphics[height=0.15\textwidth]{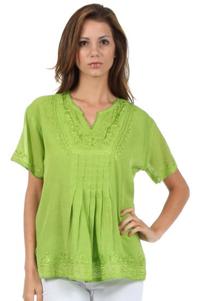}
	\includegraphics[height=0.15\textwidth]{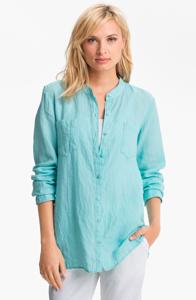}}
	\hspace{0.5cm}
	\subfloat[]{\label{ref_label5}\includegraphics[height=0.15\textwidth]{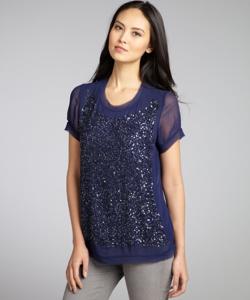}
	\includegraphics[height=0.15\textwidth]{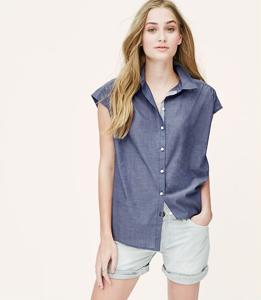}
	\includegraphics[height=0.15\textwidth]{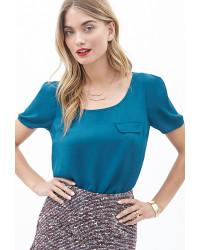}}
  \caption{\label{ref_label_overall} Some sample results demonstrating the effectiveness of the learned visual-semantic text embeddings. (a) A query image with the description \textit{``The lady was wearing a blue short sleeved blouse"}, (b)-(c) Top 3 most similar images based on their projected visual features and textual features, respectively. As can be seen, they are both semantically consistent with the input image.}
  \label{fig:inspection}
\end{figure}

\noindent\textbf{Visual-Semantic Text Embedding Evaluation}. We qualitatively evaluated the learned visual-semantic text embeddings by comparing the vector representations of the first 500 test samples with each other in a pairwise manner and inspecting the top 3 most similar images based on both their projected textual and visual features. Our analysis reveals that visual-semantic text embedding learns the relationship between images and sentences in a proper manner. Fig.~\ref{fig:inspection} shows the results of a sample query. As can be seen, the nearest neighbors in the embedding space are highly consistent with each other in terms of both modalities.\\

\begin{figure}[!t]
\centering
\begin{tabular}{c@{$\;\;\;$}c@{}c@{$\;\;$}c@{}c@{$\;\;$}c@{}c@{$\;\;$}c@{}c@{$\;\;$}c@{}c}
\scriptsize{Input} & \multicolumn{2}{c}{\scriptsize Baseline~\cite{DBLP:conf/iccv/DongYWG17}} & \multicolumn{2}{c}{\scriptsize FiLM} & \multicolumn{2}{c}{\scriptsize TV} & \multicolumn{2}{c}{\scriptsize FiLM + TV} & \multicolumn{2}{c}{\scriptsize \textbf{FiLMedGAN}}\\ 
{\includegraphics[height=1.75cm,width=1cm]{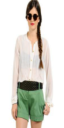}} & 
\includegraphics[height=1.75cm,width=1cm]{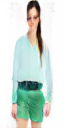} & 
\includegraphics[height=1.75cm,width=1cm]{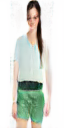} &
\includegraphics[height=1.75cm,width=1cm]{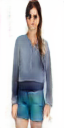} & 
\includegraphics[height=1.75cm,width=1cm]{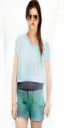} & 
\includegraphics[height=1.75cm,width=1cm]{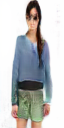} & 
\includegraphics[height=1.75cm,width=1cm]{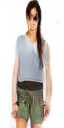} & 
\includegraphics[height=1.75cm,width=1cm]{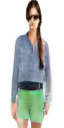} & 
\includegraphics[height=1.75cm,width=1cm]{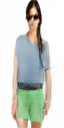} & 
\includegraphics[height=1.75cm,width=1cm]{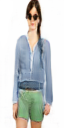} & 
\includegraphics[height=1.75cm,width=1cm]{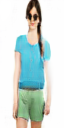} \\
{\includegraphics[height=1.75cm,width=1cm]{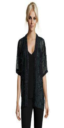}} & 
\includegraphics[height=1.75cm,width=1cm]{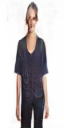} & 
\includegraphics[height=1.75cm,width=1cm]{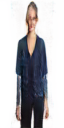} &
\includegraphics[height=1.75cm,width=1cm]{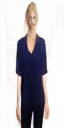} & 
\includegraphics[height=1.75cm,width=1cm]{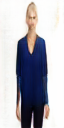} & 
\includegraphics[height=1.75cm,width=1cm]{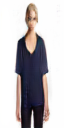} & 
\includegraphics[height=1.75cm,width=1cm]{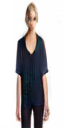} & 
\includegraphics[height=1.75cm,width=1cm]{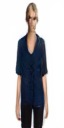} & 
\includegraphics[height=1.75cm,width=1cm]{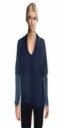} & 
\includegraphics[height=1.75cm,width=1cm]{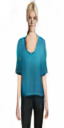} & 
\includegraphics[height=1.75cm,width=1cm]{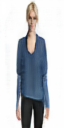}\\
{\includegraphics[height=1.75cm,width=1cm]{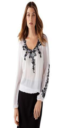}} & 
\includegraphics[height=1.75cm,width=1cm]{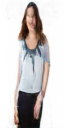} & 
\includegraphics[height=1.75cm,width=1cm]{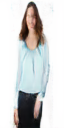} &
\includegraphics[height=1.75cm,width=1cm]{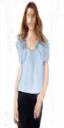} & 
\includegraphics[height=1.75cm,width=1cm]{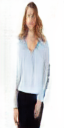} & 
\includegraphics[height=1.75cm,width=1cm]{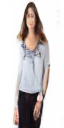} & 
\includegraphics[height=1.75cm,width=1cm]{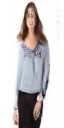} & 
\includegraphics[height=1.75cm,width=1cm]{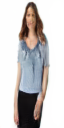} & 
\includegraphics[height=1.75cm,width=1cm]{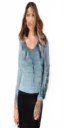} & 
\includegraphics[height=1.75cm,width=1cm]{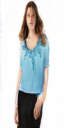} & 
\includegraphics[height=1.75cm,width=1cm]{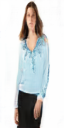}\\ 
{\includegraphics[height=1.75cm,width=1cm]{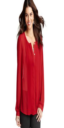}} & 
\includegraphics[height=1.75cm,width=1cm]{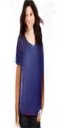} & 
\includegraphics[height=1.75cm,width=1cm]{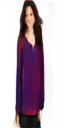} &
\includegraphics[height=1.75cm,width=1cm]{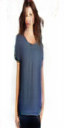} & 
\includegraphics[height=1.75cm,width=1cm]{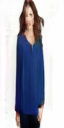} & 
\includegraphics[height=1.75cm,width=1cm]{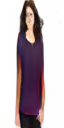} & 
\includegraphics[height=1.75cm,width=1cm]{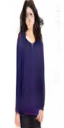} & 
\includegraphics[height=1.75cm,width=1cm]{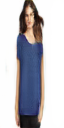} & 
\includegraphics[height=1.75cm,width=1cm]{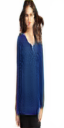} & 
\includegraphics[height=1.75cm,width=1cm]{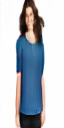} & 
\includegraphics[height=1.75cm,width=1cm]{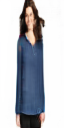}\\ 
{\includegraphics[height=1.75cm,width=1cm]{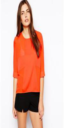}} & 
\includegraphics[height=1.75cm,width=1cm]{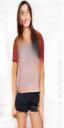} & 
\includegraphics[height=1.75cm,width=1cm]{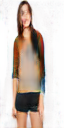} &
\includegraphics[height=1.75cm,width=1cm]{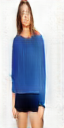} & 
\includegraphics[height=1.75cm,width=1cm]{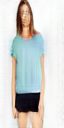} & 
\includegraphics[height=1.75cm,width=1cm]{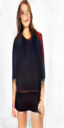} & 
\includegraphics[height=1.75cm,width=1cm]{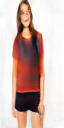} & 
\includegraphics[height=1.75cm,width=1cm]{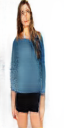} & 
\includegraphics[height=1.75cm,width=1cm]{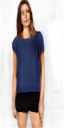} & 
\includegraphics[height=1.75cm,width=1cm]{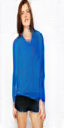} & 
\includegraphics[height=1.75cm,width=1cm]{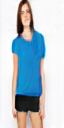}\\ 
\end{tabular}
\caption{Method comparisons. For each method, the two sample input descriptions are \textit{``the lady is wearing a blue long-sleeved blouse''} and \textit{``the lady is wearing a blue short-sleeved blouse''}, respectively.}
\label{fig:visualquality}
\end{figure}

\noindent\textbf{Qualitative Evaluation.} In Fig.~\ref{fig:visualquality}, we compare the results of our proposed FiLMedGAN model against those of the baseline method, and analyze the importance of FiLM, skip connections and TV regularization with an ablation study. While FiLM by itself exhibits a better performance in regard to language conditioned visual changes (\emph{color change in the last row}), TV regularization provides slightly more detailed images (\emph{hair and glasses in the first row}). When they are combined, the results are visually more appealing than those of the baseline model~\cite{DBLP:conf/iccv/DongYWG17}. Moreover, introducing additional skip connections, as in our FiLMedGAN model, gives the best results in terms of image details and quality since it decreases the information loss introduced by the vanilla encoder-decoder. 

For disadvantages, when the results are investigated thoroughly, it can be seen that FiLMedGAN makes the hair on the foreground disappeared while transforming the blouse. (\textit{e.g.} the rightmost image in the last row of Fig.~\ref{fig:visualquality}) Although FiLMedGAN generates plausible images in general, it may lead some degeneration on input image. It is a drawback of our approach over FashionGAN \cite{iccv2017fashiongan} where FashionGAN solves these kind of issues by using segmentation maps.

\begin{figure}[!h]
\centering
\subfloat[]{\includegraphics[height=2cm]{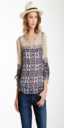}}\qquad
\subfloat[]{\includegraphics[height=2cm]{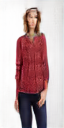}}\qquad
\subfloat[]{\includegraphics[height=2cm]{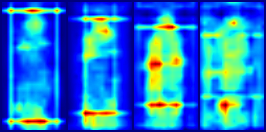}}
\caption{(a) Source image, (b) Manipulation result for the target description \textit{``the lady is wearing a red long-sleeved blouse."}, (c) Implicit attention maps of four residual FiLM blocks.}
\label{fig:filmattention}
\end{figure}

It is also important to mention that FiLM helps to visualize internal dynamics of a network and provides a kind of implicit attention mechanism. We visualize heat maps of the average filter outputs of each of four FiLMed residual layers in the generator (Fig.~\ref{fig:filmattention}). Generally speaking, while first block processes the head and the legs, third block focuses on the entire body to perform the transformations. Note that it is possible to interpret each of filter output separately rather than averaging them.\\

\begin{table}[!t]
\centering
\caption{Quantitative results. Evaluation scores are the average of last 15 epochs and the best score of the last 50 epochs is given within parenthesis.}
\begin{tabular}{l@{$\qquad$}ccccc}
\toprule
 & Baseline~\cite{DBLP:conf/iccv/DongYWG17} & FiLM & TV & FiLM+TV & \textbf{FiLMedGAN} \\
  \midrule
  IS & 2.52 (\textbf{2.68}) & 2.54 (2.65) & 2.48 (2.67) & 2.52 (2.62) & \textbf{2.58 (2.68)}\\
  FID & 22.86 (20.73) & 23.38 (20.10) & 18.79 (16.16) & 16.83 (14.84) & \textbf{10.72 (9.12)}\\
  AS & 0.65 (0.67) & 0.65 (0.66) & 0.66 (\textbf{0.68}) & 0.66 (0.67) & \textbf{0.67 (0.68)}\\
  \bottomrule
\end{tabular}
\label{table:fidscores}
\end{table}

\noindent\textbf{Quantitative Evaluation.} We apply three quantitative evaluation methods for our comparison: two for measuring realism and one for measuring the manipulation success.
For quantifying realism, we consider {Inception Score (IS)} \cite{NIPS2016_6125} and 
 {Fr\'{e}chet Inception Distance (FID)}  \cite{DBLP:journals/corr/HeuselRUNKH17}. These evaluation metrics do not measure how successful an image manipulation is done according to a target textual description, and thus we also incorporate an attribute prediction method similar to \cite{iccv2017fashiongan}. For each test image in the test set, without loss of generality, we set the next image's text description as a target text description and measure the equivalence of the modified image's attributes with the actual attributes of this next image. In order to do so, we fine-tuned a pre-trained VGG-16 \cite{DBLP:journals/corr/SimonyanZ14a} model to simultaneously predict \textit{gender}, \textit{sleeve}, \textit{color} and \textit{category} attributes. First three attributes in all likelihood could be inferred from textual description. Category attribute is thought as holistic and also included. Consequently, these attributes can be considered as representative as textual description. 

Estimated IS, FID and average attribute scores (AS) are reported in Table~\ref{table:fidscores}. In terms of AS, there is no noteworthy difference among the models through epochs. We think that this is because attributes are not distinctive enough. For example, in the first row of Fig. 3, all models have succeeded in performing the corresponding changes according to textual description and there are many such examples but the details visible in the images are open to discussion. Our {FiLMedGAN} model gives the best IS but IS is not a reliable measure~\cite{DBLP:journals/corr/abs-1801-01973}. We observed that original images have IS score $3.06$ but the early epochs of FiLM alone have $3.46$ which is absurd. It should be considered as a rough measure of quality and should not be taken seriously. According to FID, FiLM is similar to the baseline because the main role of FiLM is not to improve the quality but the conditioning. It is very interesting that, {FiLM+TV} improves the FID score in a clear way. We speculate that they contribute collaboratively to the overall result. Lastly, {FiLMedGAN} shows a significant improvement over all the other models which uses the advantage of skip connections.

\begin{comment}
  Inception & 2.5163(2.6791) & 2.5437(2.6477) & 2.4819(2.6647) & 2.5158(2.6225) & 2.5821(2.6835) \\ \hline
  AS & 0.6511(0.6697) & 0.6480(0.6645) & 0.6576(0.6754) & 0.6558(0.6663) & 0.6656(0.6781)\\ \hline
  FID & 22.8633(20.7309) & 23.3831(20.0993) & 18.7902(16.1590) & 16.8257(14.8444) & 10.7228(9.1221)\\
\end{comment}

\section{Conclusion}
We present a novel approach for language conditioned editing of fashion images. Our approach employs a GAN-based architecture which allows the users to edit an outfit image by feeding in different descriptions to generate new outfits. Our experimental analysis demonstrate that our FiLMedGAN model which employs skipping connections and FiLMed residual blocks outperforms the baselines both quantitatively and qualitatively and generates more plausible outfit images according to a given natural description.

\bibliographystyle{splncs04}
\bibliography{egbib}

\begin{thebibliography}{10}
\providecommand{\url}[1]{\texttt{#1}}
\providecommand{\urlprefix}{URL }
\providecommand{\doi}[1]{https://doi.org/#1}

\bibitem{DBLP:journals/corr/abs-1801-01973}
Barratt, S., Sharma, R.: A note on the inception score. CoRR
  \textbf{abs/1801.01973} (2018), \url{http://arxiv.org/abs/1801.01973}

\bibitem{pmlr-v70-bergmann17a}
Bergmann, U., Jetchev, N., Vollgraf, R.: Learning texture manifolds with the
  periodic spatial {GAN}. In: Proceedings of the 34th International Conference
  on Machine Learning (ICML) (2017)

\bibitem{bojanowski2017enriching}
Bojanowski, P., Grave, E., Joulin, A., Mikolov, T.: Enriching word vectors with
  subword information. Transactions of the Association for Computational
  Linguistics (TACL)  (2017)

\bibitem{chen2017language}
Chen, J., Shen, Y., Gao, J., Liu, J., Liu, X.: Language-based image editing
  with recurrent attentive models. CoRR  \textbf{abs/1711.06288} (2017),
  \url{http://arxiv.org/abs/1711.06288}

\bibitem{Chen2018AttentionGANFO}
Chen, X., Xu, C., Yang, X., Tao, D.: Attention-gan for object transfiguration
  in wild images. CoRR  \textbf{abs/1803.06798} (2018),
  \url{http://arxiv.org/abs/1803.06798}

\bibitem{cho-al-emnlp14}
Cho, K., van Merri{\"{e}}nboer, B., G{\"{u}}l{\c c}ehre, {\c C}., Bahdanau, D.,
  Bougares, F., Schwenk, H., Bengio, Y.: Learning phrase representations using
  rnn encoder--decoder for statistical machine translation. In: Proceedings of
  the 2014 Conference on Empirical Methods in Natural Language Processing
  (EMNLP) (2014)

\bibitem{DBLP:conf/iccv/DongYWG17}
Dong, H., Yu, S., Wu, C., Guo, Y.: Semantic image synthesis via adversarial
  learning. In: Proceedings of the IEEE International Conference on Computer
  Vision (ICCV) (2017)

\bibitem{vunet2018}
Esser, P., Sutter, E., Ommer, B.: A variational u-net for conditional
  appearance and shape generation. In: Proceedings of IEEE Conference on
  Computer Vision and Pattern Recognition (CVPR) (2018)

\bibitem{NIPS2014_5423}
Goodfellow, I., Pouget-Abadie, J., Mirza, M., Xu, B., Warde-Farley, D., Ozair,
  S., Courville, A., Bengio, Y.: Generative adversarial nets. In: Advances in
  Neural Information Processing Systems 27 (NIPS) (2014)

\bibitem{han2017viton}
Han, X., Wu, Z., Wu, Z., Yu, R., Davis, L.S.: Viton: An image-based virtual
  try-on network. In: Proceedings of IEEE Conference on Computer Vision and
  Pattern Recognition (CVPR) (2018)

\bibitem{DBLP:journals/corr/HeuselRUNKH17}
Heusel, M., Ramsauer, H., Unterthiner, T., Nessler, B., Klambauer, G.,
  Hochreiter, S.: Gans trained by a two time-scale update rule converge to a
  {N}ash equilibrium. CoRR  \textbf{abs/1706.08500} (2017),
  \url{http://arxiv.org/abs/1706.08500}

\bibitem{pmlr-v37-ioffe15}
Ioffe, S., Szegedy, C.: Batch normalization: Accelerating deep network training
  by reducing internal covariate shift. In: Proceedings of the 32nd
  International Conference on Machine Learning (ICML) (2015)

\bibitem{DBLP:conf/iccvw/JetchevB17}
Jetchev, N., Bergmann, U.: The conditional analogy {GAN:} swapping fashion
  articles on people images. In: 2017 IEEE International Conference on Computer
  Vision Workshops (ICCVW) (2017)

\bibitem{DBLP:journals/corr/JetchevBV16}
Jetchev, N., Bergmann, U., Vollgraf, R.: Texture synthesis with spatial
  generative adversarial networks. CoRR  \textbf{abs/1611.08207} (2016),
  \url{http://arxiv.org/abs/1611.08207}

\bibitem{Johnson2016Perceptual}
Johnson, J., Alahi, A., Fei-Fei, L.: Perceptual losses for real-time style
  transfer and super-resolution. In: European Conference on Computer Vision
  (ECCV) (2016)

\bibitem{Kastaniotis2018AttentionAwareGA}
Kastaniotis, D., Ntinou, I., Tsourounis, D., Economou, G., Fotopoulos, S.:
  Attention-aware generative adversarial networks (ata-gans). CoRR
  \textbf{abs/1802.09070} (2018), \url{http://arxiv.org/abs/1802.09070}

\bibitem{DBLP:journals/corr/KingmaB14}
Kingma, D.P., Ba, J.: Adam: {A} method for stochastic optimization. In: The
  International Conference on Learning Representations (ICLR) (2015)

\bibitem{Lassner:GeneratingPeople:2017}
Lassner, C., Pons-Moll, G., Gehler, P.V.: A generative model for people in
  clothing. In: Proceedings of the IEEE International Conference on Computer
  Vision (ICCV) (2017)

\bibitem{liu2016deepfashion}
Liu, Z., Luo, P., Qiu, S., Wang, X., Tang, X.: Deepfashion: Powering robust
  clothes recognition and retrieval with rich annotations. In: Proceedings of
  IEEE Conference on Computer Vision and Pattern Recognition (CVPR) (2016)

\bibitem{NIPS2017_6644}
Ma, L., Jia, X., Sun, Q., Schiele, B., Tuytelaars, T., Van~Gool, L.: Pose
  guided person image generation. In: Advances in Neural Information Processing
  Systems 30 (NIPS) (2017)

\bibitem{ma2018disentangled}
Ma, L., Sun, Q., Georgoulis, S., Van~Gool, L., Schiele, B., Fritz, M.:
  Disentangled person image generation. In: Proceedings of IEEE Conference on
  Computer Vision and Pattern Recognition (CVPR) (2018)

\bibitem{DBLP:journals/corr/abs-1802-06454}
Ma, S., Fu, J., Chen, C.W., Mei, T.: {DA-GAN:} instance-level image translation
  by deep attention generative adversarial networks (with supplementary
  materials). CoRR  \textbf{abs/1802.06454} (2018),
  \url{http://arxiv.org/abs/1802.06454}

\bibitem{Mahendran2015}
Mahendran, A., Vedaldi, A.: Understanding deep image representations by
  inverting. In: Proceedings of IEEE Conference on Computer Vision and Pattern
  Recognition (CVPR) (2015)

\bibitem{N18-2120}
Manjunatha, V., Iyyer, M., Boyd-Graber, J., Davis, L.: Learning to color from
  language. In: Proceedings of the 2018 Conference of the North American
  Chapter of the Association for Computational Linguistics: Human Language
  Technologies, Volume 2 (Short Papers) (2018)

\bibitem{4756141}
Nilsback, M., Zisserman, A.: Automated flower classification over a large
  number of classes. In: Indian Conference on Computer Vision, Graphics and
  Image Processing (ICVGIP) (2008)

\bibitem{perez:hal-01648685}
Perez, E., Strub, F., De~Vries, H., Dumoulin, V., Courville, A.: {FiLM: Visual
  Reasoning with a General Conditioning Layer}. In: {AAAI Conference on
  Artificial Intelligence}. New Orleans, United States (Feb 2018),
  \url{https://hal.inria.fr/hal-01648685}

\bibitem{pumarola2018unsupervised}
Pumarola, A., Agudo, A., Sanfeliu, A., Moreno-Noguer, F.: {Unsupervised Person
  Image Synthesis in Arbitrary Poses}. In: Proceedings of IEEE Conference on
  Computer Vision and Pattern Recognition (CVPR) (2018)

\bibitem{pmlr-v48-reed16}
Reed, S., Akata, Z., Yan, X., Logeswaran, L., Schiele, B., Lee, H.: Generative
  adversarial text to image synthesis. In: Proceedings of The 33rd
  International Conference on Machine Learning (ICML) (2016)

\bibitem{RFB15a}
Ronneberger, O., P.Fischer, Brox, T.: U-net: Convolutional networks for
  biomedical image segmentation. In: Medical Image Computing and
  Computer-Assisted Intervention (MICCAI). LNCS, vol.~9351, pp. 234--241.
  Springer (2015),
  \url{http://lmb.informatik.uni-freiburg.de/Publications/2015/RFB15a},
  (available on arXiv:1505.04597 [cs.CV])

\bibitem{Rudin1992}
Rudin, L.I., Osher, S., Fatemi, E.: Nonlinear total variation based noise
  removal algorithms. Physica D: Nonlinear Phenomena  \textbf{60}(1-4),
  259--268 (nov 1992). \doi{10.1016/0167-2789(92)90242-f},
  \url{https://doi.org/10.1016/0167-2789(92)90242-f}

\bibitem{Salimans:2016:ITT:3157096.3157346}
Salimans, T., Goodfellow, I., Zaremba, W., Cheung, V., Radford, A., Chen, X.:
  Improved techniques for training gans. In: Advances in Neural Information
  Processing Systems 29 (NIPS) (2016)

\bibitem{NIPS2016_6125}
Salimans, T., Goodfellow, I., Zaremba, W., Cheung, V., Radford, A., Chen, X.,
  Chen, X.: Improved techniques for training gans. In: Advances in Neural
  Information Processing Systems 29 (NIPS) (2016)

\bibitem{siarohin:hal-01761539}
Siarohin, A., Sangineto, E., Lathuili{\`e}re, S., Sebe, N.: {Deformable GANs
  for Pose-based Human Image Generation}. In: Proceedings of IEEE Conference on
  Computer Vision and Pattern Recognition (CVPR) (2018)

\bibitem{DBLP:journals/corr/SimonyanZ14a}
Simonyan, K., Zisserman, A.: Very deep convolutional networks for large-scale
  image recognition. CoRR  \textbf{abs/1409.1556} (2014),
  \url{http://arxiv.org/abs/1409.1556}

\bibitem{WahCUB_200_2011}
Wah, C., Branson, S., Welinder, P., Perona, P., Belongie, S.: {The Caltech-UCSD
  Birds-200-2011 Dataset}. Tech. Rep. CNS-TR-2011-001, California Institute of
  Technology (2011)

\bibitem{Xian_2018_CVPR}
Xian, W., Sangkloy, P., Agrawal, V., Raj, A., Lu, J., Fang, C., Yu, F., Hays,
  J.: Texturegan: Controlling deep image synthesis with texture patches. In:
  Proceedings of IEEE Conference on Computer Vision and Pattern Recognition
  (CVPR) (2018)

\bibitem{wang2018sftgan}
Xintao~Wang, Ke~Yu, C.D., Loy, C.C.: Recovering realistic texture in image
  super-resolution by deep spatial feature transform. In: Proceedings of IEEE
  Conference on Computer Vision and Pattern Recognition (CVPR) (2018)

\bibitem{Tao18attngan}
Xu, T., Zhang, P., Huang, Q., Zhang, H., Gan, Z., Huang, X., He, X.: Attngan:
  Fine-grained text to image generation with attentional generative adversarial
  networks. In: Proceedings of IEEE Conference on Computer Vision and Pattern
  Recognition (CVPR) (2018)

\bibitem{Zanfir_2018_CVPR}
Zanfir, M., Popa, A.I., Zanfir, A., Sminchisescu, C.: Human appearance
  transfer. In: Proceedings of IEEE Conference on Computer Vision and Pattern
  Recognition (CVPR) (2018)

\bibitem{DBLP:journals/corr/abs-1805-08318}
Zhang, H., Goodfellow, I.J., Metaxas, D.N., Odena, A.: Self-attention
  generative adversarial networks. CoRR  \textbf{abs/1805.08318} (2018),
  \url{http://arxiv.org/abs/1805.08318}

\bibitem{Han17stackgan2}
Zhang, H., Xu, T., Li, H., Zhang, S., Wang, X., Huang, X., Metaxas, D.N.:
  Stackgan++: Realistic image synthesis with stacked generative adversarial
  networks. CoRR  \textbf{abs/1710.10916} (2017),
  \url{http://arxiv.org/abs/1710.10916}

\bibitem{Zhang_2017_ICCV}
Zhang, H., Xu, T., Li, H., Zhang, S., Wang, X., Huang, X., Metaxas, D.N.:
  Stackgan: Text to photo-realistic image synthesis with stacked generative
  adversarial networks. In: Proceedings of the IEEE International Conference on
  Computer Vision (ICCV) (2017)

\bibitem{zhang2018hdgan}
Zhang, Z., Xie, Y., Yang, L.: Photographic text-to-image synthesis with a
  hierarchically-nested adversarial network. In: Proceedings of IEEE Conference
  on Computer Vision and Pattern Recognition (CVPR) (2018)

\bibitem{iccv2017fashiongan}
Zhu, S., Fidler, S., Urtasun, R., Lin, D., Loy, C.C.: Be your own {P}rada:
  Fashion synthesis with structural coherence. In: Proceedings of the IEEE
  International Conference on Computer Vision (ICCV) (2017)

\end{thebibliography}
\end{document}